# A Two-Level Clinical Novelty Prioritization Framework for Disease Trajectories Discovered from Longitudinal Electronic Health Records

Jin-Jia Zheng[1], Vincent S. Tseng[1,*]

[1] Department of Computer Science, National Yang Ming Chiao Tung University, Taiwan

* Corresponding author: vtseng@cs.nycu.edu.tw

**Abstract**

Graph-based analysis of electronic health records (EHRs) yields disease-trajectory catalogs, but structural support does not establish semantic novelty. We present a two-level framework to prioritize 99,527 branching motifs previously discovered by an upstream patient-disjoint, train-only MIMIC-IV v3.1 pipeline and frozen after validation-based catalog selection. The framework separates path-level and node-to-endpoint semantic sparsity from reliability-adjusted clinical priority. Three deterministic textualizations per motif or node-target pair were embedded with BAAI/bge-small-en-v1.5 and calibrated against 166,972 pre-2023 PubMed references using mean top-20 cosine distance. Positive recombination occurred in 8.63% of motifs and had negligible effect on global ranking, whereas specificity and support/GNN reliability terms changed clinical-priority ordering. Robustness analyses across neighbor counts, templates, encoders, generic-label taxonomies, score settings, PrimeKG, and corrected density-based RND showed that broad ordering was more stable than hard top-k membership. A semantic-dose analysis did not support a monotonic relation between score and forecasting benefit after endpoint and token-count adjustment (1.96 days per SD; 95% CI, -0.75 to 4.68). In the frozen catalog-q90 high-semantic subgroup, branch-enhanced forecasting reduced conditional event-time MAE by 20.1 days (95% CI, 13.3-27.0) and sMAPE by 2.37 percentage points (95% CI, 1.81-2.93), while discrimination and endpoint-ranking differences were inconclusive. Corrected RND showed low tail agreement with the primary score (Spearman rho = 0.430; q90 Jaccard = 0.125) despite a similar aggregate timing association. These findings support representation-aware, uncertainty-aware literature-relative prioritization of discovered trajectories rather than causal discovery or clinician-adjudicated novelty.

## 1. Introduction

Longitudinal electronic health records (EHRs) reveal how diagnoses accumulate, branch, and precede clinical endpoints. Resources such as MIMIC-IV [1, 2] support longitudinal informatics [3], while prior studies have examined population disease trajectories [4] and patient representation learning [5-11]. Graph-based analyses have already produced inspectable paths and branching motifs. This study does not seek to discover new trajectories; it asks which previously discovered candidates warrant literature review, model inspection, or expert assessment.

Support and graph confidence are reliability signals rather than definitions of novelty. Conversely, literature-based discovery, atypical-combination analyses, semantic predication resources, and embedding-based novelty methods provide complementary ways to characterize underexplored concepts, relations, or documents [12-16]. However, a single full-path score cannot reveal whether novelty arises from uncommon node-endpoint relations or from an unusual combination of otherwise familiar relations. Broad or administrative labels may also inflate semantic distance without increasing clinical informativeness.

We address this downstream task by prioritizing previously discovered trajectories rather than introducing another discovery method. Path-level novelty evaluates a complete motif, node-level novelty summarizes its node-to-endpoint relations, and positive recombination captures the positive path-minus-node difference. A separate clinical-priority score applies a generic-label penalty and support/GNN reliability bonuses. These scores rank retrospective hypotheses; they are not probabilities of clinical novelty.

The input trajectories were discovered from MIMIC-IV v3.1 for four endpoints: chronic obstructive pulmonary disease (COPD), stroke, chronic kidney disease (CKD), and dementia. Upstream graph evidence and candidate motifs used train-only data; catalog policies were selected on validation data and frozen before test evaluation. The semantic scoring settings were also fixed without test outcomes. This study contributes (1) a two-level literature-calibrated semantic representation of already-discovered motifs, (2) separate semantic and reliability-adjusted priority scores, (3) final-catalog component, generic-taxonomy, and textualization-uncertainty analyses, (4) robustness checks across neighbor counts, encoders, templates, weights, endpoint composition, and ontology paths, and (5) censor-aware downstream validation with patient-cluster uncertainty and multiplicity sensitivity.

## 2. Related work

### 2.1 Disease trajectories and sequential EHR modeling

Longitudinal EHR research has developed along two closely related directions: explicit disease-trajectory analysis and predictive sequence representation learning. Population trajectory studies convert repeated diagnoses into temporally ordered disease relations and have shown that large-scale health records can reveal recurring progression patterns and clinically meaningful sequences [4]. In parallel, Deep Patient, Doctor AI, Med2Vec, RETAIN, Dipole, BEHRT, and Med-BERT demonstrate that patient histories can be encoded into representations that support future-event prediction, risk stratification, and diagnosis forecasting [5-11]. These models differ in architecture and training objective, but all exploit the premise that the ordering and context of prior events contain information about future clinical states.

The two directions offer complementary strengths. Explicit trajectory mining yields inspectable sequences and branches that can be reviewed as candidate disease-progression hypotheses, whereas neural sequence models typically compress history into latent states or contextualized token representations [4-11]. The latter are optimized primarily for predictive utility, not for deciding whether a discovered progression pattern is already well established in biomedical knowledge. Conversely, a trajectory that is frequent or predictive in an EHR does not automatically constitute a novel clinical observation. This distinction becomes especially important when discovery produces tens of thousands of candidate motifs rather than a small manually curated list.

Trajectory discovery and novelty prioritization are different stages. Here, the branching-motif catalog was already discovered and frozen through upstream structural criteria and validation-based selection. We do not generate new disease trajectories. Instead, we rank the existing motifs by literature-relative uncommonness, specificity, and graph reliability. This separates predictive performance and structural support from semantic novelty while preserving explicit motifs for expert review.

### 2.2 Graph mining and reliability

Graph-based clinical modeling represents diagnoses, relations, or patients as nodes connected by structured edges. GRAM demonstrates how graph structure can regularize medical-code representations using hierarchical knowledge, while graph convolutional and graph attention architectures provide general mechanisms for aggregating local neighborhood information [17-19]. In disease-trajectory discovery, analogous graph principles can be applied to empirical diagnosis-transition networks, where edge support, transition confidence, learned link scores, and branching structure describe how strongly a candidate pattern is supported by observed longitudinal data.

These quantities are useful measures of discovery reliability, but they should not be interpreted as novelty. A high-support transition can be common precisely because it represents a well-known clinical relationship, whereas a low-frequency pattern can be either genuinely underexplored or simply noisy. Likewise, a high GNN link score indicates compatibility with learned graph structure, not semantic distance from the biomedical literature. The distinction is methodologically important because combining reliability and novelty too early can make it impossible to determine whether a high final score reflects uncommonness, structural plausibility, or both.

Knowledge graphs provide a related but different source of structured evidence. Resources such as PrimeKG organize curated biomedical entities and relations that can support ontology- or path-based comparisons [20]. Such graphs are valuable for testing whether semantically unusual EHR trajectories are also structurally distant in an external knowledge space, but incomplete code mapping and differences between observational EHR transitions and curated biomedical relations limit their use as a gold standard. Accordingly, our framework reports pure semantic novelty separately from reliability-adjusted clinical priority and treats graph-derived measures as supporting evidence rather than definitions of novelty.

### 2.3 Literature-based discovery and semantic novelty

Literature-based discovery (LBD) was developed to identify meaningful connections that are distributed across otherwise disconnected parts of the biomedical literature. Swanson's early work showed that complementary literatures could imply previously unrecognized hypotheses, and later interactive LBD systems formalized the search for bridging concepts between document sets [12, 13]. This tradition is highly relevant to EHR trajectory analysis because a disease path may be composed of individually familiar diagnoses whose particular combination or progression has received limited explicit attention.

Scientific-novelty research has also examined atypical combinations as a signal of potentially novel work [14]. In biomedicine, SemMedDB provides a large-scale repository of semantic predications extracted from PubMed and can support relation-oriented knowledge discovery [15]. More recently, SemNovel uses embedding distance to estimate the semantic novelty of biomedical publications relative to a reference corpus [16]. These approaches move beyond literal keyword overlap, allowing semantically related documents or concepts to be compared even when they are expressed with different terminology.

Applying these ideas to EHR-derived trajectories introduces additional representational choices that prior publication-level novelty measures do not directly address. A branching motif contains both local node-to-endpoint relations and a whole-path configuration; novelty may arise because individual relations are uncommon, because familiar relations are recombined in an unusual way, or both. In addition, novelty estimates depend on how a trajectory is textualized, which reference documents are available, the embedding model, and the number of nearest neighbors used for calibration. Our two-level design therefore separates node-level and path-level evidence, explicitly defines positive recombination, and audits sensitivity to wording, encoder, and neighbor count rather than treating a single embedding-distance value as ground truth.

### 2.4 Biomedical representation learning and prioritization

Modern semantic novelty analysis depends strongly on the representation model used to encode clinical and biomedical text. The Transformer architecture provides the foundation for contextual token representations [21], while BERT introduced large-scale bidirectional pretraining for transferable language understanding [22]. Biomedical and clinical variants such as BioBERT, ClinicalBERT, and domain-specific biomedical pretraining adapt this paradigm to scientific articles and clinical text [23-25]. Self-alignment methods for biomedical entities further improve concept-level representations, and large multimodal clinical concept embeddings show that distributed representations can recover meaningful medical relationships from heterogeneous data sources [26, 27].

These models make semantic comparison feasible at scale, but they also introduce representation dependence. Different encoders may emphasize lexical similarity, biomedical entity identity, contextual relations, or broader topical semantics. For trajectory novelty, deterministic textualization adds another layer of uncertainty because the same underlying motif can be phrased in multiple reasonable ways. A high ranking that is stable only under one template or one encoder is therefore weaker evidence than a ranking that remains broadly consistent across independently specified representations. This motivates our use of multiple deterministic wording variants, an alternative encoder, and template-family sensitivity analyses.

Finally, semantic uncommonness is not identical to clinical priority. Broad administrative or non-specific labels can appear distant from the literature corpus for reasons that have little to do with clinically meaningful novelty, while structurally well-supported motifs may deserve review even when their semantic distance is moderate. We therefore treat prioritization as a multi-factor problem: pure semantic evidence is reported first, then a transparent specificity correction and graph-reliability terms are added to obtain a separate clinical-priority score. This separation preserves interpretability of the underlying novelty signal and supports uncertainty-aware ranking for downstream case review, external validation, and forecasting analyses rather than automated claims of discovery.

## 3. Methods

### 3.1 Framework overview

The input is a frozen catalog of endpoint-specific branching motifs previously discovered by an upstream train-only graph pipeline. Our framework operates on these existing motifs without rerunning discovery or using novelty to select the catalog. Each motif is decomposed into node-target pairs and a full-path representation, then textualized and calibrated against an external biomedical reference distribution. Pure semantic scores are reported separately from clinical-priority scores adjusted for specificity and graph reliability. Explicit paths illustrate the general decomposition, but the primary analysis uses only the 99,527 previously discovered branching motifs.

Fig. 1 summarizes the prioritization of previously discovered disease trajectories, from decomposition and semantic scoring of the frozen motif catalog to reliability-adjusted clinical priority.

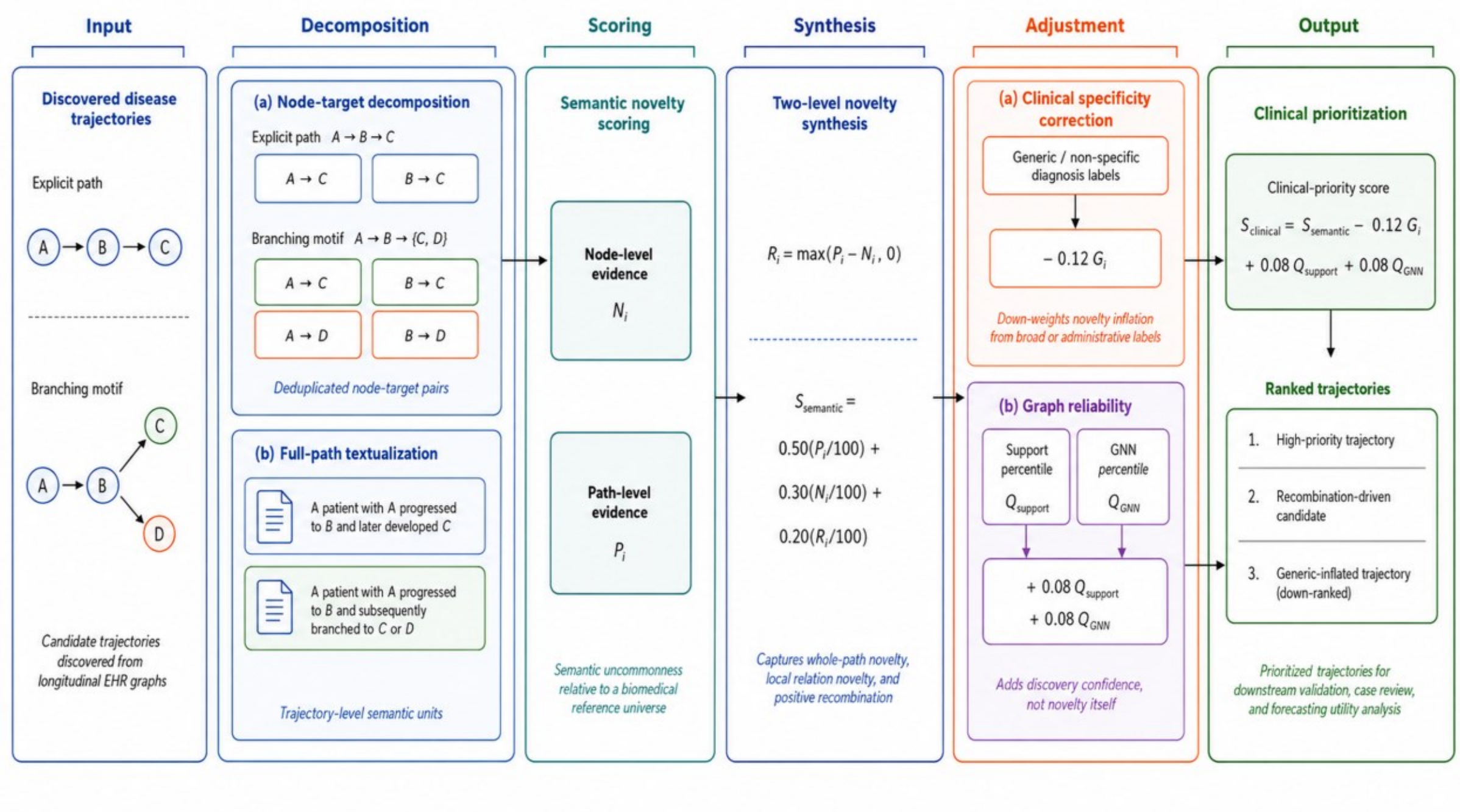

Fig. 1. Two-level clinical novelty prioritization of previously discovered disease trajectories. The frozen branching-motif catalog from an upstream train-only discovery stage is decomposed into deterministic node-target and full-path textualizations, calibrated against external literature distances, and combined into separate pure semantic and reliability-adjusted priority scores. The explicit-path schematic illustrates the general decomposition; only previously discovered branching motifs are used in the primary analysis.

### 3.2 Cohort, leakage control, and frozen motif catalog

We used MIMIC-IV v3.1, accessed on 29 July 2025 [1, 2]. ICD-9-CM diagnoses were mapped to CCS and ICD-10-CM diagnoses to CCSR. All mapped primary and secondary diagnoses per admission were retained and deduplicated; XXX000 and CCS259 were excluded. Patients required at least two retained admissions. COPD mapped to RSP008/CCS127, stroke to CIR020/CIR021/CIR022/CIR024/CIR025, CKD to GEN003/CCS158, and dementia to NVS011.

In the upstream discovery stage, 100,018 patients were assigned by subject_id to disjoint training, validation, and test sets (70:10:20; seed 20260714) before graph construction. The training graph included 749 diagnosis categories. A two-layer GraphSAGE-style link predictor used six structural node features, 128 hidden units, 64-dimensional embeddings, and a directed edge MLP. One-to-many and prefix-to-many motifs were ranked using support, transition confidence, GNN evidence, and branch entropy. Validation AUPRC selected prespecified stable/broad policies without a fixed top-N export cap. This catalog was frozen before the novelty analyses.

Table 1. Frozen cohort and endpoint-adaptive motif catalog.

| Item | Train | Validation | Test | Total / setting |
|---|---|---|---|---|
| Patients | 70,012 | 10,001 | 20,005 | 100,018 |
| Admissions | 296,433 | 41,580 | 84,083 | 422,096 |
| Raw index examples | 226,421 | 31,579 | 64,078 | 322,078 |
| Eligible 3,650-day indices | 56,054 | 7,441 | 15,676 | 79,171 |
| Catalog policy | - | Selected by AUPRC | Frozen | No top-N cap |
| COPD motifs | - | - | - | 22,897 (stable) |
| Stroke motifs | - | - | - | 15,828 (broad) |
| CKD motifs | - | - | - | 42,377 (broad) |
| Dementia motifs | - | - | - | 18,425 (stable) |
| All motifs | - | - | - | 99,527 |

### 3.3 Semantic artifact and deterministic textualization

The frozen semantic artifact contained 99,527 motifs, 298,581 path textualizations (three per motif), 16,667 unique node-target pairs, and 50,001 node textualizations (three per pair). Textualizations were generated by fixed templates (endpoint_branch_semnovel_v2 and node_target_novelty_v1), not by an online generative model.

Text was embedded using BAAI/bge-small-en-v1.5 [28] at revision 5c38ec7c405ec4b44b94cc5a9bb96e735b38267a. The 384-dimensional CLS embedding was L2-normalized. The reference universe comprised 166,972 aligned PubMed Title/Abstract records published before 2023. For query q, raw novelty was the mean cosine distance to the 20 nearest reference embeddings:

$$d(q) = (1/20)\ \Sigma_{j=1}^{20}\ [1 - \cos(e_q, e_j)].$$

A sample of 1,000 external reference records formed an empirical non-self reference-to-reference distance distribution. Query percentile P(q) was the percentage of calibration distances not exceeding d(q). The final percentile for a motif or node-target pair was the mean over its three wording variants. Thus, an external 95th percentile indicates sparsity relative to the external reference-distance distribution, not membership in the top 5% of the motif catalog.

### 3.4 Two-level semantic novelty

For motif i, P_i is the full-path external percentile. N_i is the mean external percentile of its source/intermediate node-to-endpoint pairs. Positive recombination is R_i = max(P_i - N_i, 0). Percentiles are divided by 100 before aggregation:

$$S_i\text{semantic} = 0.50(P_i/100) + 0.30(N_i/100) + 0.20(R_i/100).$$

The weights were prespecified heuristics and were not learned or selected using test forecasting performance. Recombination is a derived contrast, not an independent semantic source.

### 3.5 Clinical-priority score and generic-label audit

$$S_i clinical = S_i semantic - 0.12G_i + 0.08Q_i support + 0.08Q_i GNN.$$

G_i is the fraction of parsed trajectory labels matching a case-normalized description rule for broad labels such as other/other specified, unspecified, not elsewhere classified, abnormal findings, aftercare, encounter, screening, status, social administration, evaluation, and signs or symptoms. Qsupport and QGNN are full-catalog percentile ranks in [0,1]. The rule is a transparent heuristic, not a clinical ontology or adjudicated specificity standard. The primary penalty is 0.12; 0.48 is retained only as a post-hoc sensitivity setting.

### 3.6 Threshold systems and downstream subgroup

External path/node quadrants use P_i >= 95 and N_i >= 95. They characterize literature-relative semantic sparsity. Downstream high-semantic membership instead uses the frozen global q90 of the pure semantic score, Ssemantic >= 0.7843067. For each patient-index-endpoint row, semantic evidence is the maximum Ssemantic among at most 24 retrieved endpoint motifs. Because a row is high when any retrieved motif exceeds q90, the high-row fraction need not equal 10%. The clinical-priority q90 (0.862813) is not used for this primary subgroup. Dose-response analyses additionally used frozen motif-catalog quintiles and q50/q70/q80/q90/q95 cuts after row-level maximum aggregation; ties and missing-score rows were retained, and one patient could contribute to different strata across indices or endpoints.

### 3.7 Censor-aware downstream evaluation

Each nonterminal admission was a potential index. Endpoints observed at or before the index were prevalent and removed from the at-risk set. Events within 30 days were washout exclusions. For horizon H = 3,650 days, an endpoint was positive if first observed from day 30 through H, negative if known not to occur through H, and unknown when no event was observed but ascertainment follow-up was shorter than H. Unknown rows were masked from loss and evaluation. Follow-up ended at the last recorded hospital discharge or an earlier death.

The final test evaluation included 15,676 index examples from 4,631 patients, 62,704 endpoint rows, 27,122 valid at-risk rows, and 17,066 observed positive events. Ranking excluded prevalent, washout, right-censor-unknown, and other non-at-risk endpoints. MRR used the first ranked eligible positive endpoint; 73.56% of 13,630 ranking examples had only one at-risk candidate, creating a ceiling effect.

The branch-enhanced model was a two-layer time-aware Transformer with endpoint-specific cross-attention over branching-motif tokens; it did not contain a novelty gate, learned novelty-aware selector, or novelty auxiliary loss. Novelty therefore defines a post hoc, frozen evidence subgroup rather than changing the final forecasting architecture.

### 3.8 Statistical analysis

Forecasting values are means across model-initialization seeds 42-46. Paired 95% intervals used 5,000 unique-patient cluster bootstrap resamples for the primary subgroup analyses; each sampled patient contributed all repeated index admissions and endpoint rows, with identical draws for both models and all seeds. Textualization uncertainty used 1,000 resamples within the three wording variants of each motif and node-target pair while holding the patient cohort and literature corpus fixed. Dose-response subgroup intervals used 1,000 paired patient resamples, and the continuous dose association used patient-cluster-robust regression of branch-minus-history absolute-error difference on standardized semantic score with endpoint and retrieved-token-count adjustment. Component-sensitivity intervals used endpoint-stratified motif resampling. These uncertainty analyses answer different questions and were not pooled. Holm adjustment was applied only within named comparison families; percentile CIs are pointwise, and bootstrap-tail quantities are not exact permutation-test p-values.

### 3.9 Semantic robustness and ontology comparison

Robustness analyses used the identical frozen 99,527-motif catalog and did not use test outcomes to select semantic settings. The primary nearest-neighbor count remained k = 20, while semantic scores were recomputed for k in {1, 5, 10, 20, 50, 100}. Representation sensitivities included BAAI/llm-embedder (revision c3e8ac86ff4859012570d6b59341fe67b88e55ad; 768-dimensional CLS embeddings, L2-normalized), PubMedBERT-based embeddings [25], two BGE deterministic text families (neutral_sequence and structured_fields), 1,000/10,000/20,000 calibration samples, a progression-focused literature corpus, generic-label taxonomies, and 313 prespecified score/component settings. Each independent representation fit its own catalog-level global q90 before downstream evaluation; thresholds were never estimated from test-row quantiles, and the primary BGE k = 20 representation remained frozen.

A density-adjusted RND sensitivity used a frozen 20,000-document exact-text-deduplicated pre-2023 reference sample. For each query, a local-density log ratio compared mean k-nearest-reference cosine distance (k=20) with the neighbors' leave-one-out local reference distance. Path, node, and recombination quantities were mapped to 0-100 percentiles and combined on the same normalized scale as the primary score: 0.50(path/100) + 0.30(node/100) + 0.20(recombination/100). Cosine near-duplicate thresholds of 0.99 and 0.98 were evaluated as sensitivities. No EHR test label entered RND scoring or catalog-threshold fitting.

As a structured knowledge-source comparison, catalog diagnosis labels were mapped to the Precision Medicine Knowledge Graph (PrimeKG) [20], and shortest-path distance was treated as a coarse ontology-path baseline. Because only 118 of 522 catalog codes had accepted mappings and full predecessor mapping was uncommon, this analysis was interpreted as a robustness comparison rather than a gold standard for clinical novelty.

### 3.10 Blinded review protocol and external-validation status

An endpoint-balanced blinded review packet containing 80 cases (40 matched high-versus-low pairs) and a locked analysis plan were prepared with algorithmic scores and sampling strata concealed. No clinician ratings were used in the present analysis. Independent external-EHR validation was not performed because no approved compatible longitudinal EHR and diagnosis mapping were available.

## 4. Results

### 4.1 Final catalog distribution and external quadrants

Across 99,527 motifs, mean path and node external percentiles were 92.16 and 96.29, respectively; the node distribution was especially concentrated. This indicates a domain/template mismatch between fixed clinical-trajectory wording and the external literature calibration. External percentiles should therefore be interpreted as semantic sparsity in the chosen embedding space, not probabilities of true clinical novelty.

Table 2. External-95th-percentile quadrants in the frozen catalog.

| Quadrant | Motifs | Fraction | Mean $G_i$ | Any generic |
|---|---|---|---|---|
| Path-high / node-high | 36,690 | 36.86% | 0.2152 | 91.38% |
| Path-high / node-low | 1,030 | 1.03% | 0.1466 | 82.82% |
| Path-low / node-high | 40,632 | 40.83% | 0.1790 | 87.87% |
| Path-low / node-low | 21,175 | 21.27% | 0.1355 | 77.71% |
| Total | 99,527 | 100.00% | 0.1828 | 86.95% |

### 4.2 Generic taxonomy, recombination, and component sensitivity

The current description-based taxonomy marked 118 of 513 diagnosis descriptions as generic. Across the frozen catalog, 86.95% of motifs contained at least one matched generic label, whereas the mean per-motif generic-label fraction was 0.1828; these quantities are not interchangeable. Narrow and broad taxonomy variants marked 25 and 134 labels, respectively. Narrow versus current ranking had Spearman rho = 0.9598 and top-10% Jaccard = 0.5484, whereas current versus broad yielded rho = 0.9923 and Jaccard = 0.8517. Changing the generic penalty from 0.12 to 0.48 produced rho = 0.7703, mean absolute rank change 15,037, and top-300 overlap 0.763. Because 0.48 was post hoc, the prespecified 0.12 remains primary.

Positive recombination occurred in 8,588 of 99,527 motifs (8.63%), with a mean weighted contribution of 0.000144 to the semantic score. Removing recombination barely changed global order (rho = 0.999861). In a clinical-priority q90 sensitivity, removing the recombination term yielded row-membership Jaccard 0.989404 (358 rows left the high group and 21 entered); the branch-minus-history MAE difference was -22.428 days under the original score and -22.675 days without recombination. The pointwise timing intervals were similar, but the scoped family adjustment did not establish a difference. Across 313 weight/component settings, 81.8% retained rho >= 0.90 and 62.6% retained top-10% Jaccard >= 0.70 relative to the primary clinical-priority ranking. A four-axis Pareto analysis identified 230 motifs (0.23%) on the non-dominated front. Recombination is therefore retained as a diagnostic contrast rather than evidence of synergistic clinical novelty.

Table 3. Final-catalog component and weight sensitivity.

| Variant | Spearman vs. primary | 95% motif-resampling CI | Top-300 overlap |
|---|---|---|---|
| Semantic-only | 0.417700 | [0.393485, 0.442123] | 0.003 |
| Reliability-only | 0.733372 | - | 0.077 |
| No reliability bonuses | 0.498498 | [0.475079, 0.520692] | 0.057 |
| Without recombination | 0.999861 | [0.999821, 0.999891] | 0.987 |
| Node-only | 0.319679 | - | 0.000 |
| Path-only | 0.421102 | - | 0.003 |
| Generic penalty 0.48 | 0.770333 | - | 0.763 |

### 4.3 Textualization-evidence uncertainty

Hard global top-k membership was only moderately stable under resampling of deterministic wording variants. Stability improved at larger k and differed substantially by endpoint. The point-estimate q90 set contained 9,954 motifs; 6,782 remained high in at least 80% of resamples and 5,653 in at least 95%, indicating a stable core despite unstable exact ordering.

Table 4. Textualization-evidence bootstrap ranking stability.

| Scope | Top-k | Mean overlap | 95% evidence interval |
|---|---|---|---|
| Global | 100 | 0.521 | [0.450, 0.590] |
| Global | 300 | 0.592 | [0.553, 0.633] |
| Global | 1,000 | 0.679 | [0.660, 0.699] |
| CKD | 100 | 0.537 | [0.470, 0.600] |
| COPD | 100 | 0.711 | [0.650, 0.770] |
| Dementia | 100 | 0.657 | [0.590, 0.720] |
| Stroke | 100 | 0.842 | [0.790, 0.890] |

### 4.4 Frozen high-semantic subgroup

Semantic evidence was available for 58,328 endpoint rows; 35,538 rows met the frozen global q90 criterion. Among at-risk high-semantic rows, 9,255 positive events were observed. The subgroup was enriched for stroke and dementia, whereas the non-high subgroup contained more COPD and CKD events. Performance comparisons must therefore be interpreted with endpoint composition and difficulty in mind.

Table 5. Five-seed downstream performance by frozen semantic subgroup.

| Subgroup | Model | AUROC | AUPRC | MRR | Hit@1 | Gap MAE | sMAPE |
|---|---|---|---|---|---|---|---|
| High q90 | History-only | 0.859678 | 0.912640 | 0.904746 | 0.824085 | 669.40 | 67.72% |
| High q90 | Branch-enhanced | 0.861536 | 0.913948 | 0.903381 | 0.821283 | 649.26 | 65.34% |
| Non-high | History-only | 0.878370 | 0.958903 | 0.973137 | 0.947323 | 685.28 | 79.66% |
| Non-high | Branch-enhanced | 0.888677 | 0.962555 | 0.978054 | 0.957339 | 661.07 | 76.48% |

Table 6. Paired patient-cluster bootstrap differences in the high-semantic subgroup.

| Metric | Branch-enhanced improvement | 95% CI | Interpretation |
|---|---|---|---|
| AUROC | +0.001858 | [-0.000792, 0.004500] | Not supported |
| AUPRC | +0.001308 | [-0.000493, 0.003219] | Not supported |
| MRR | -0.001365 | [-0.003760, 0.000976] | Not supported |
| Hit@1 | -0.002802 | [-0.007246, 0.001570] | Not supported |
| Gap MAE | 20.14 days lower | [13.27, 26.98] | Timing improved |
| Gap sMAPE | 2.37 points lower | [1.81, 2.93] | Timing improved |

#### 4.4.1 Semantic dose-response and missing-score analysis

Catalog-score quintiles contained 345, 969, 2,495, 9,528, and 44,991 endpoint rows (212, 485, 1,092, 2,459, and 4,561 unique patients) and 164, 373, 797, 3,133, and 12,345 observed positive rows, respectively. Branch-minus-history MAE differences were -17.885, -12.016, -30.270, -23.097, and -21.501 days for Q1-Q5. A missing-score group contained 4,376 endpoint rows, 1,631 patients, and 254 positives and showed +162.328 days (95% CI 94.345-228.545), indicating substantially worse branch timing in that conditional subgroup; this is not a token-removal or causal effect. Q90 and Q95 differences were -20.138 and -19.158 days. After adjusting for endpoint and retrieved-token count, a one-SD increase in semantic score had coefficient +1.961 days for branch-minus-history absolute-error difference (95% cluster-robust CI -0.754 to 4.677; 3,764 patients, 16,812 positive rows). Thus, no monotonic semantic-dose benefit was established. Across the 39-comparison dose family, the smallest Holm-adjusted bootstrap-tail quantity was approximately 0.0779 at 1,000-resample resolution.

### 4.5 Endpoint-specific high-semantic results

Within the primary high-semantic subgroup, COPD showed discrimination gains (AUROC +0.01374, 95% CI 0.00619-0.02142; AUPRC +0.01013, 95% CI 0.00501-0.01579) and an MAE reduction of 11.16 days (95% CI 4.53-18.02). CKD, dementia, and stroke discrimination intervals crossed zero, while their MAE reductions were 16.40, 21.25, and 25.81 days, respectively, with unadjusted intervals excluding zero. These endpoint analyses were exploratory. In the endpoint-by-metric multiplicity sensitivity family, Holm adjustment retained COPD AUROC, AUPRC, MAE, and sMAPE; CKD and dementia MAE and sMAPE; and stroke sMAPE. The stroke MAE comparison did not remain supported after the primary family-wise correction.

Because the high-semantic subgroup was compositionally enriched for stroke and dementia, timing effects were also recomputed under endpoint-composition controls. For the primary BGE representation, MAE reductions were 18.16 days (95% CI 12.08-24.10) under the common natural mix, 18.66 days (12.53-24.77) under an equal-endpoint meta-average, and 18.50 days (11.45-25.37) under endpoint-count matching; corresponding sMAPE reductions were 2.21, 2.24, and 2.28 percentage points. Composition-standardized timing improvement was therefore broadly retained, although sensitivity was representation-dependent; for example, the neutral-template count-matched MAE interval crossed zero.

### 4.6 Encoder, textualization, and neighbor-count robustness

All four independent representation reruns used the same 99,527 motifs, pre-2023 reference universe, three textualizations per item, and frozen five-seed forecasting predictions. Broad motif ordering was moderately to strongly preserved, but exact extreme membership was representation-sensitive. Patient-row agreement with the primary high-q90 definition was incomplete, and alternative templates changed endpoint composition. These shifts limit pooled subgroup interpretation and motivate representation-specific frozen thresholds.

Table 7. Motif-level and downstream robustness across semantic representations (Jacc. = Jaccard overlap).

| Representation | Motif rho | Top-300 overlap | q90 Jacc. | High rows | MAE reduction, d (95% CI) | sMAPE reduction, points (95% CI) |
|---|---|---|---|---|---|---|
| Primary BGE | Ref. | Ref. | 1.000 | 35,538 | 20.14 [13.27, 26.98] | 2.37 [1.81, 2.93] |
| Second encoder | 0.7464 | 0.047 | 0.335 | 28,515 | 23.21 [14.01, 32.33] | 2.49 [1.78, 3.21] |
| Neutral template | 0.8580 | 0.047 | 0.387 | 27,803 | 21.04 [11.80, 29.92] | 2.42 [1.69, 3.15] |
| Structured fields | 0.7899 | 0.267 | 0.424 | 24,513 | 22.91 [13.28, 32.35] | 2.59 [1.83, 3.35] |

Neighbor-count sensitivity showed the same pattern. Across k = 1, 5, 10, 20, 50, and 100, broad ordering remained correlated but hard top sets changed: k = 1 versus k = 20 yielded rho = 0.9387 and top-100 Jaccard = 0.3889, whereas k = 1 versus k = 100 yielded rho = 0.8350 and top-100 Jaccard = 0.1765. High-semantic subgroup MAE reductions ranged from 19.08 to 22.84 days across all six k values, with confidence intervals excluding zero; AUPRC improvement had clear support only at k = 1, and MRR was neutral or slightly negative. The primary k = 20 was retained because it was prespecified rather than selected from downstream test performance.

#### 4.6.1 Corrected density comparator and representation dependence

After normalization of all RND composite components to the common 0-100 percentile scale, the corrected RND score had Spearman rho = 0.430072 and catalog-q90 Jaccard = 0.125445 versus the primary score, demonstrating strong method dependence of exact tail membership. The corrected RND high-q90 subgroup contained 36,365 endpoint rows, 4,505 unique patients, and 10,576 positive rows; branch-minus-history MAE was -25.230 days (95% CI -32.462 to -18.234). Near-duplicate thresholds of cosine 0.99 and 0.98 produced nearly identical aggregate differences (-25.229 and -25.204 days). High-group AUPRC intervals crossed zero, and the non-high group also showed timing improvement, so these findings do not establish novelty-specific effect modification.

Other representation checks showed the same distinction between broad order and exact tail membership. A progression-focused literature corpus had Spearman rho = 0.950851 and q90 Jaccard = 0.665300 versus the primary composite, whereas PubMedBERT yielded 0.576470 and 0.181311. Motif-level external-95 saturation was 37.899% for path and 77.697% for node percentiles under the primary representation, but 99.477% and 100% under the structured template family. Similar aggregate downstream timing associations therefore do not imply encoder-invariant motif identities or clinical novelty.

### 4.7 External-threshold and ontology-path sensitivity

External-reference 95th-percentile rules and catalog q90 captured different constructs. Among 58,328 rows with semantic scores, catalog q90 classified 60.9% as high, whereas external path95, node95, both95, and either95 classified 93.3%, 99.6%, 92.8%, and 99.7%, respectively. External-both95 retained an MAE reduction of 21.99 days (95% CI 16.09-27.85) and an sMAPE reduction of 2.65 points (95% CI 2.14-3.17), but its 92.8% coverage approached the full semantically available cohort. We therefore retained frozen catalog q90 for the primary downstream subgroup and used external 95th-percentile rules only for quadrant interpretation and distribution-shift sensitivity.

The PrimeKG shortest-path baseline captured a different and relatively coarse signal. Accepted mappings were available for 118 of 522 catalog codes; 98.83% of motifs had at least one mapped predecessor, but only 1.42% had every predecessor mapped. Sentence-based semantic novelty and ontology distance had Spearman rho = 0.1557, with top-100 Jaccard = 0.0753 and q90-set Jaccard = 0.0797. The ontology-q90 subgroup showed a 15.28-day MAE reduction (95% CI 7.90-22.06), compared with 24.63 days (18.20-31.10) outside the subgroup; AUPRC improved by 0.00123 (0.00054-0.00198), whereas MRR and Hit@1 intervals crossed zero. PrimeKG therefore provides a complementary robustness baseline, not an ontology gold standard or clinical validation.

### 4.8 Blinded review and external-validation status

The endpoint-balanced 80-case blinded review packet and locked analysis plan were completed, but qualified clinicians had not yet submitted the required independent ratings. Accordingly, no clinician agreement, score-correlation, or clinician-validation result is reported. External-EHR validation was also not conducted because no approved compatible longitudinal EHR and diagnosis mapping were available.

### 4.9 Anonymous model-behavior cases

Four anonymized seed-42 cases were selected by the largest positive risk-probability decrease after deletion of the top motif within each endpoint's eligible high-semantic observed events. They illustrate computational sensitivity rather than independent clinical validation. The CKD example is deliberately retained despite a large timing error.

Table 8. Anonymous high-semantic model-behavior cases.

| Case | Endpoint | Semantic max | Full risk | Baseline risk | Observed gap | Predicted gap | Top-1 deletion drop |
|---|---|---|---|---|---|---|---|
| HN_CASE_001 | COPD | 0.7866 | 0.8385 | 0.6982 | 338.0 d | 1292.9 d | 0.0582 |
| HN_CASE_002 | Stroke | 0.7871 | 0.9144 | 0.8254 | 1542.2 d | 1977.7 d | 0.1287 |
| HN_CASE_003 | CKD | 0.7915 | 0.3313 | 0.5119 | 3468.2 d | 250.7 d | 0.0970 |
| HN_CASE_004 | Dementia | 0.7970 | 0.6507 | 0.5161 | 1521.0 d | 2192.8 d | 0.0233 |

## 5. Discussion

The results refine the role of two-level novelty. Path and node evidence remain conceptually complementary semantic units, but positive recombination was uncommon (8.63%), had negligible influence on global rank, and produced nearly identical q90 membership and downstream timing when removed. In contrast, generic-label specificity and support/GNN reliability materially changed cross-endpoint clinical-priority ordering. Ssemantic should therefore be reported as pure literature-relative evidence and Sclinical as a reliability-adjusted prioritization score; recombination is best treated as a diagnostic contrast rather than evidence of synergistic novelty.

The textualization bootstrap and neighbor-count analysis show why a deterministic global top-k list should not be treated as fixed. Exact top-100 membership was only moderately stable under wording resampling, and increasing semantic-neighbor count changed hard top sets more than broad rank order. Nevertheless, a large core of catalog-q90 motifs remained high across wording resamples, and event-time improvement persisted across k = 1-100. Endpoint-specific inclusion frequencies, score components, rank intervals, and endpoint composition are therefore more informative than a single cross-endpoint hard ranking.

Semantic-dose analyses further narrowed the downstream claim. Timing improvements appeared across multiple catalog strata rather than increasing monotonically with semantic score, and the endpoint/token-adjusted continuous coefficient crossed zero. The missing-score subgroup showed marked timing degradation, but this is a conditional association and not evidence that deleting semantic evidence causes harm. Because the dose family contained 39 comparisons and used 1,000 resamples, pointwise intervals and finite-resolution family adjustments should be interpreted separately.

Encoder, corpus, template, and density-based analyses sharpen this conclusion. Broad semantic ordering could remain correlated while extreme motif identities, patient-row membership, and endpoint composition changed appreciably. The corrected RND comparator had low agreement with the primary tail (rho 0.430; q90 Jaccard 0.125), and PubMedBERT was also substantially less concordant than the progression-focused corpus. Representation-specific high-score groups nevertheless showed similar aggregate conditional timing associations, so stability of downstream averages should not be confused with invariance of motif identity or a clinically validated novelty construct.

The external-threshold comparison further exposed EHR-to-literature distribution shift. Maximum aggregation over retrieved motifs caused external 95th-percentile rules to label nearly all semantically available rows as high, and motif-level percentile saturation varied sharply with textualization. Their positive timing results are therefore closer to broad-cohort sensitivity than selective novelty validation. PrimeKG shortest-path distance agreed only weakly with sentence-based novelty and had incomplete code mapping, indicating that the representations capture different signals. These findings support catalog-relative q90 as the primary downstream semantic subgroup and treat external percentiles, density scores, alternative corpora, and ontology distance as complementary sensitivities rather than ground truth.

Downstream validation produced a narrower conclusion than the historical selector experiment. The final forecasting model did not use novelty during training or token selection. Within the frozen high-semantic subgroup, branch tokens reduced conditional event-time error, but overall discrimination and endpoint-ranking differences were inconclusive. The non-high subgroup also showed timing and discrimination improvements, semantic-dose analyses were non-monotonic, and the high subgroup was enriched for stroke and dementia. The results therefore do not support a claim that the branch-enhanced model is preferentially effective for clinically novel disease trajectories; they support a downstream temporal association with literature-relative semantic strata.

A supporting repair analysis of the upstream graph-discovery model provides an important qualification for the GNN reliability term. Directed incoming/outgoing message passing with target-degree-matched negatives substantially improved patient-disjoint unseen-edge discovery relative to the original GNN. However, when the 99,527 previously discovered motif identities and binary token availability were held fixed, substituting the repaired GNN-derived scores did not improve discrimination or ranking and increased timing MAE by 2.10 days (95% CI 0.35-3.86 days worse). Stronger transition prediction therefore did not automatically yield stronger forecasting, and GNN reliability should be interpreted as one structural prioritization axis rather than a validated proxy for downstream utility. Attention and motif-deletion analyses likewise provide computational sensitivity only, not causal or clinician-validated explanations.

## 6. Limitations

First, MIMIC-IV represents a single hospital system. Censor-aware masking prevents observed insufficient follow-up from being labeled negative, but the last recorded discharge cannot guarantee endpoint-free follow-up elsewhere. Second, CCS/CCSR aggregation can merge heterogeneous diagnoses, and the generic-label taxonomy is a description-based heuristic rather than a clinician-adjudicated ontology; narrow, broad, and ontology-derived alternatives changed hard top-set membership. Third, semantic novelty depends on the literature universe, encoder, deterministic textualization, neighbor count, local-density definition, and endpoint composition. Corrected RND, PubMedBERT, alternative templates, corpus sensitivity, and k-sensitivity preserved aggregate timing associations more consistently than exact high-ranked motif identities. External percentiles were concentrated in the upper tail and could saturate under some templates, consistent with EHR-to-literature representation mismatch.

Fourth, global clinical-priority ranking is sensitive to endpoint composition and reliability terms, whereas recombination contributed little under the final setting. Fifth, semantic-dose strata were highly unequal, some low-dose strata were nearly all positive, and the adjusted continuous association did not support monotonic effect modification; unsupported AUC/AP cells were retained as not applicable. Sixth, the exploratory dose family used 1,000 bootstrap resamples, limiting the resolution of tail quantities, and reported percentile CIs remain pointwise rather than simultaneous. Seventh, downstream masked endpoint ranking is ceiling-limited because most evaluated indices contain only one at-risk candidate. Eighth, the repaired GNN improved unseen-edge discovery but did not improve fixed-catalog forecasting scores. Finally, no external-EHR validation, completed blinded clinician adjudication, or human semantic calibration is available. The method is intended for retrospective hypothesis prioritization, not diagnosis, treatment selection, causal inference, or automated claims of clinical discovery.

## 7. Conclusion

We developed a leakage-controlled two-level semantic novelty framework to prioritize a previously discovered and frozen catalog of 99,527 branching motifs. The method begins after graph-based disease-trajectory discovery and separates path- and node-level literature-relative evidence from reliability-adjusted clinical priority. It quantifies sensitivity to generic-label taxonomy, score components, wording, semantic-neighbor count, encoder, reference corpus, local-density scoring, endpoint composition, and ontology paths. Reliability and specificity terms materially affected global clinical-priority ordering, whereas recombination had little effect and semantic-dose analyses did not establish a monotonic relationship with forecasting benefit. Broad rankings and aggregate conditional timing associations were more robust than exact hard top-k identities; corrected RND showed low tail overlap with the primary score despite a similar timing association. In the frozen high-semantic subgroup, branch-enhanced forecasting reduced conditional event-time error, but discrimination and endpoint-ranking gains were inconclusive and similar benefits also occurred outside the subgroup. These results support uncertainty-aware, representation-aware retrospective prioritization of already-discovered trajectories while limiting claims about clinical novelty, causality, clinician validation, and cross-institution generalization.